\documentclass{article}

\usepackage{arxiv}

\usepackage[utf8]{inputenc} 
\usepackage[T1]{fontenc}    
\usepackage{hyperref}       
\usepackage{url}            
\usepackage{booktabs}       
\usepackage{amsfonts}       
\usepackage{nicefrac}       
\usepackage{microtype}      
\usepackage{lipsum}
\usepackage{graphicx}
\usepackage{adjustbox,makecell,color,amsmath}

\title{Multimodal MRI Accurately Identifies Amyloid Status in Unbalanced Cohorts in \\Alzheimer's Disease Continuum}

\author{Giorgio Dolci$^{1,2,3}$, Charles A. Ellis$^{3}$, Federica Cruciani$^2$, Lorenza Brusini$^2$, Anees Abrol$^3$, \\ Ilaria Boscolo Galazzo$^2$, Gloria Menegaz$^{2,+}$, and Vince D. Calhoun$^{3,+}$,\\ for the Alzheimer’s Disease Neuroimaging Initiative$^*$ \\\\
$^1$Department of Computer Science, University of Verona, Verona, Italy\\
$^2$Department of Engineering for Innovation Medicine, University of Verona, Verona, Italy\\
$^3$Tri-Institutional Center for Translational Research in Neuroimaging and Data Science (TReNDS),\\ Georgia State University, Georgia Institute of Technology, Emory University, Atlanta, GA, USA\\\\
$^+$V.D. Calhoun and G. Menegaz equally contributed as last authors to this work.\\
$^*$Data used in preparation of this article were obtained from the Alzheimer’s Disease \\Neuroimaging Initiative (ADNI) database (adni.loni.usc.edu). As such, the investigators within the ADNI \\contributed to the design and implementation of ADNI and/or provided data but did not participate in analysis \\or writing of this report. A complete listing of ADNI investigators can be found at: \\\url{http://adni.loni.usc.edu/wp-content/uploads/how_to_apply/ADNI_Acknowledgement_List.pdf}}

\begin{document}
\maketitle
\begin{abstract}
Amyloid-$\beta$ (A$\beta$) plaques in conjunction with hyperphosphorylated tau proteins in the form of neurofibrillary tangles are the two neuropathological hallmarks of Alzheimer’s disease.
It is well-known that the identification of individuals with A$\beta$ positivity could enable early diagnosis.
In this work, we aim at capturing the A$\beta$ positivity status in an unbalanced cohort enclosing subjects at different disease stages, exploiting the underlying structural and connectivity disease-induced modulations as revealed by structural, functional, and diffusion MRI. Of note, due to the unbalanced cohort, the outcomes may be guided by those factors rather than amyloid accumulation. The partial views provided by each modality are integrated in the model allowing to take full advantage of their complementarity in encoding the effects of the A$\beta$ accumulation, leading to an accuracy of $0.762\pm0.04$.
The specificity of the information brought by each modality is assessed by \textit{post-hoc} explainability analysis (guided backpropagation), highlighting the underlying structural and functional changes.
Noteworthy, well-established biomarker key regions related to A$\beta$ deposition could be identified by all modalities, including the hippocampus, thalamus, precuneus, and cingulate gyrus, witnessing in favor of the reliability of the method as well as its potential in shading light on modality-specific possibly unknown A$\beta$ deposition signatures.
\end{abstract}


\section{Introduction}

The amyloid cascade hypothesis in Alzheimer's Disease (AD) posits that the accumulation of extracellular Amyloid-$\beta$ (A$\beta$) neuritic plaques in the brain leads to tauopathy and consequent neurodegeneration (\cite{haass2022if}). Hence, A$\beta$ deposition in the brain is considered to be the first step and the principal trigger of AD pathology (\cite{fernandez2023cortical,haass2022if}). In consequence,  neurodegeneration of gray matter (GM) may be related to the deposition of A$\beta$ plaques, resulting in cerebral atrophy and synaptic loss (\cite{serrano2011neuropathological}), altering many brain regions, especially subcortical areas, and functional networks. Additionally, A$\beta$ plaques in AD are also linked with demyelination of white matter (WM) tracts (\cite{sanchez2020white}).

In clinical practice, the precise identification of A$\beta$ proteins, and thus the classification of patients as A$\beta$ positive or negative, is performed through amyloid positron emission tomography (PET) scans and cerebrospinal fluid (CSF) tests. Despite their undeniable utility, their use has some inherent limitations. First, limiting to one modality hides informative features that would require other kinds of investigation tools to be revealed.
Then, amyloid PET uses ionizing radiation and radioactive tracers, is expensive, and is not widely available (\cite{hansson2018csf, lee2021performance}). Additionally, the CSF test requires an invasive lumbar puncture.

A wider perspective on the disease can be obtained by adding views obtained by other imaging modalities. Among these, non-invasive MRI techniques are at the top of the list, allowing to assess structural and functional changes encoding the effect of the amyloid accumulation that can profitably be used as additional biomarkers.
Structural MRI (sMRI) has been widely used for AD detection and early prediction due to its ability to detect GM atrophy and structural changes. Resting-state functional MRI (rs-fMRI) detects changes in blood oxygenation level-dependent (BOLD) signals, which depend on neurovascular coupling, and hence indirectly measures brain neural activity (\cite{johnson2012brain}). Different resting-state networks (RSNs), like the saliency and default-mode (DM) networks, have been shown to be altered in AD pathology (\cite{pini2021breakdown,palmqvist2017earliest,zhou2010divergent}). Finally, diffusion MRI (dMRI) is an imaging technique that relies on the movement of water molecules, allowing the assessment of both microstructural (\cite{zucchelli2016lies}) and structural connectivity (SC) changes (\cite{sanchez2020white}).
These two modalities are generally used to map the whole-brain connectomes, with rs-fMRI describing the functional connectivity (FC) between region pairs (usually in terms of correlations) and dMRI the SC (most often relying on the number of WM fibers linking the target regions).

Different studies focusing on A$\beta$ classification tasks (positive vs negative) in AD research have employed PET images and related features with deep learning (DL) models, while only a few works used MRI data. In \cite{son2020clinical}, a slice-level approach for the identification of the A$\beta$ status was considered in conjunction with a 2D-CNN for feature extraction and classification. In the same manner, \cite{reith2020application} adopted the slice-level approach for the classification of A$\beta$ status employing two different 2D-CNN (ResNet-50 and ResNet-152), both reaching high accuracy (around $0.95$).
On the other hand, \cite{lee2021performance} considered three different well-known 3D-CNNs for detecting A$\beta$ positivity relying on 3D Florbetaben brain PET images. \cite{kim2021deep} developed a particular CNN composed of different submodules for analyzing the 3D FDG-PET images, converting them into slices following the three different axes. \cite{kang2023alzheimer} faced the problem of A$\beta$ classification relying on both early and delay-phase FBB PET images and tested their models considering both single and fused modalities. Using FDG and Amyloid PET-derived images, they were able to achieve competitive performance in this task, with accuracies of around $0.80$. More recently, \cite{rasi2024predicting} employed FDG-PET-derived features in order to predict the A$\beta$ status in the AD continuum. To this end, they tested eight different feature selection methods and eight different classifiers. LASSO in conjunction with the Gaussian Naive Bayes (GNB) model performed better with respect to the others, achieving an AUC of $0.924$.
Regarding MRI-based approaches for A$\beta$ detection, \cite{chattopadhyay2023predicting} employed sMRI images in conjunction with a 3D-CNN, while \cite{yang2021deep} used an SDF-based convolutional network to analyze the hippocampus region. Using sMRI-derived images and features, they were able to reach accuracies around $0.75$.
Due to the heterogeneous factors that lead to AD, in recent years, many studies have focused on multimodal DL models due to their ability to integrate information of different nature and to outperform single-modality methods (\cite{dolci2023deep, chattopadhyay2023predicting, abrol2019multimodal}). 
Recently, graph neural and convolutional networks (GNNs/GCNs) have become popular in neuroscience due to their perfect fit for functional and structural brain networks. \cite{cui2022braingb} proposed a benchmark for analyzing fMRI and dMRI networks through GNNs, testing different messages passing, node features, and pooling operations, while \cite{wee2019cortical} developed a GCN to study the cortical thickness.

Although DL models can achieve high performance in different tasks, they do not easily provide interpretable output for what they have learned, which is particularly problematic in clinical and biomedical domains. 
To address this issue, eXplainable Artificial Intelligence (XAI) methods have been developed allowing to identify the contributions of input features to final predictions, potentially highlighting crucial information for AD (\cite{abrol2020deep, abrol2021deep, bohle2019layer, el2021multilayer}).

In this study, the overall goal of the proposed approach consists of decrypting the signatures induced by A$\beta$ accumulation in the views provided by s/rs-f/dMRI, taking advantage of their complementarity for capturing the A$\beta$ status while exploiting their interplay for the identification of the regions that played a prominent role in determining the outcome. 
To instantiate this idea, we present a multimodal and explainable DL framework for the classification of A$\beta$ positive and negative status relying on an unbalanced cohort of individuals spanning the AD continuum. The proposed framework includes structural and functional connectomes derived from dMRI and rs-fMRI, respectively, along with sMRI-derived GM 3D volumes in order to investigate complementary aspects as well as their relations. For the sake of interpretability of the results, an extensive \textit{post-hoc} XAI analysis was then performed, pointing the spotlight on the input features that most influenced the final outcome.

\section{Materials and Methods}

\subsection{Dataset}
    The sMRI, rs-fMRI, and dMRI neuroimaging data used in the preparation of this article were obtained from the Alzheimer’s Disease Neuroimaging Initiative (ADNI) database (\url{adni.loni.usc.edu}). The ADNI was launched in 2003 as a public-private partnership, led by Principal Investigator Michael W. Weiner, MD. The primary goal of ADNI has been to test whether serial MRI, PET, other biological markers, and clinical and neuropsychological assessment could be combined to measure the progression of mild cognitive impairment (MCI) and early AD. For up-to-date information, please refer to \url{www.adni-info.org}.

    One of the key strengths of the dataset is the inclusion/exclusion criteria adopted to recruit the subjects; subjects with neurological diseases other than AD and with different substance/drug use were excluded from the study. Due to this, the results we uncover related to A$\beta$ are unlikely to be related to other diseases or the use of a particular substance.
    For more information about exclusion criteria, please refer to the official document at this link: \url{https://adni.loni.usc.edu/wp-content/themes/freshnews-dev-v2/documents/consentForms/ADNI3_ProtocolVersion3.1_20201204.pdf}.

    In this work, our dataset was initially composed of $18416$ preprocessed sMRI images (from $2144$ subjects) from the ADNI $1$, $2$, $3$, and GO phases, out of which $18334$ passed quality control (QC) (from $2143$ subjects). For rs-fMRI, $2584$ preprocessed images were considered (from $1143$ individuals) from the ADNI $2$, $3$, and GO phases, out of which $2450$ passed QC (from $1105$ individuals). For dMRI, $901$ preprocessed images (from $901$ subjects) from ADNI $3$ were considered, out of which $894$ (from $894$ subjects) passed QC. Additional information about QC is detailed in the next paragraph. The images of the first available timepoint from only subjects belonging to control (CN), significant memory concern (SMC), early MCI (EMCI), late MCI (LMCI), and AD clinical classes, that had all three modalities, and available A$\beta$ status were included. Lumbar puncture to retrieve CSF samples was performed using the procedures described on the ADNI website, and subjects were labeled as A$\beta$ positive or negative based on the A$\beta$ protein levels reported by the CSF test. Similarly to \cite{hansson2018csf}, a cutoff of $980 \text{pg/mL}$ was used to define the A$\beta$ status (i.e., $<980 \text{pg/mL}$ for positivity).

    Aiming at the classification based on the A$\beta$ status as the target outcome, the considered group of individuals was further split gathering the A$\beta$ negative CN, SMC, and EMCI individuals in the NEG class ($69$, $75$, and $41$ subjects), and the A$\beta$ positive EMCI, LMCI, and AD in the POS class ($53$, $53$, and $27$ subjects), respectively, resulting in an unbalanced data split with respect to the disease stage. The LMCI and AD groups were not included in the A$\beta$ negative class since they could represent different underlying conditions linked with the functional decline (e.g., Lewy Body dementia, Frontotemporal dementia, vascular dementia, TDP-43 pathology). In contrast, CN and SMC individuals with amyloid accumulation are subjects of ongoing debate, with no clear consensus on whether they represent a prodromal stage of AD or individuals at higher risk of developing the disease but still may never develop the disease (\cite{jack2020preclinical,frisoni2019re}).
    
    \begin{table}[!ht]
        \centering
        \begin{adjustbox}{width=0.95\linewidth}
        \begin{tabular}{c c c c c c c c c c c c}
            \toprule
            \makecell{Status} & \makecell{\# of subjects} & \makecell{Age} & \makecell{Sex (M/F)} & \makecell{MMSE} & \makecell{A$\beta_{42}$} & \makecell{} & \makecell{CN} & \makecell{SMC} & \makecell{EMCI} & \makecell{LMCI} & \makecell{AD}\\
            \midrule
            \makecell{A$\beta$ -} & \makecell{$185$} & \makecell{$71.8\pm7.1$} & \makecell{$68/117$} & \makecell{$28.9\pm1.5$} & \makecell{$1684.3\pm601.2$} & \makecell{} & \makecell{$69$} & \makecell{$75$} & \makecell{$41$} & \makecell{-} & \makecell{-}\\
            \makecell{A$\beta$ +} & \makecell{$133$} & \makecell{$74.5\pm7.5$} & \makecell{$71/62$} & \makecell{$25.3\pm4.3$} & \makecell{$607.5\pm189.3$} & \makecell{} & \makecell{-} & \makecell{-} & \makecell{$53$} & \makecell{$53$} & \makecell{$27$}\\
            \bottomrule
        \end{tabular}
        \end{adjustbox}
        \caption{Demographic information of the A$\beta$ cohort patients.}
        \label{tab:demos_info}
    \end{table}

    The MRI images for the considered cohort were collected as follows: i) T1-weighted sMRI: TE/TR=shortest, TI=$900$ ms, FOV=$256\times256$ mm$^2$, $1$ mm isotropic resolution, slices=$176-211$; ii) rs-fMRI: TE/TR=$30/3000$ ms, FOV=$220\times220\times163$ mm$^3$, $3.4$ mm isotropic resolution, $200$ volumes in almost all subjects, with minimal variations (e.g., $195-197$) in a small subset; iii) single-shell dMRI: TE/TR=$56/7200$ ms, FOV=$232\times232\times160$ mm$^3$, $2$ mm isotropic resolution, b=$0$ and $1000$ s/mm$^2$. 

    \subsection{Preprocessing and Feature extraction}

    The sMRI preprocessing included tissue segmentation of GM, WM, and CSF with the modulated normalization algorithm in the statistical parametric mapping toolbox (SPM12, \url{http://www.fil.ion.ucl.ac.uk/spm/}). This work used GM volumes smoothed with a Gaussian kernel (FWHM=$6$mm). For QC, images that had a low correlation with individual and/or group-level masks were discarded, which involved correlating data at three levels: the entire image, the top $20$ slices, and the bottom $20$ slices. The full preprocessed GM volume was input to the sMRI channel of the neural network, resulting in an input size of $121\times145\times121$ for each subject.

    Adhering to the process proposed in \cite{du2020neuromark}, the rs-fMRI data was preprocessed with SPM12 including rigid body motion correction, removal of scans with high head motion parameters ($>3^\circ$ of rotations and $>3mm$ in translations), slice-timing correction, warping to the standard MNI space using the EPI template, resampling to ${3mm}^3$ isotropic voxels, and smoothing with a Gaussian kernel (FWHM = $6$ mm). QC was the same as for sMRI, correlating the data at three levels: the entire image, the top $5$ slices, and the bottom $5$ slices. Fifty-three maximally independent components (ICs) covering the whole brain were extracted using spatially constrained ICA with the Neuromark\_fMRI\_$1.0$ template (available in the GIFT software; \url{http://trendscenter.org/software/gift}). The ICs were divided into $7$ RSNs: the i) Sub-cortical (SuC); ii) Auditory (AU); iii) Sensorimotor (SM); iv) Visual (VI); v) Cognitive-control (CC); vi) DM); and, vii) Cerebellar (CB) networks. For each subject, the Pearson correlation between IC time courses was computed, resulting in a $53$x$53$ static functional network connectivity (FNC) matrix, where FNC is the network analog of FC in that the timecourses represent weighted partially overlapping whole brain patterns. 
    Finally, each FNC matrix was converted into a complete, undirected, and weighted graph. The edges' weights correspond to the FNC correlation values, considering both positive and negative values, while the values of the $53$ nodes (i.e., ICs) were initialized at a value of one in order to force the network to learn a latent representation based only on the connectivity information. This FNC-based graph was the input to the rs-fMRI channel.
    
    The dMRI volumes were preprocessed via brain extraction followed by Eddy currents correction (FSL 6.0, \url{https://fsl.fmrib.ox.ac.uk/}). The data was then denoised using local principal component analysis (PCA) via empirical thresholds relying on the Python \textit{dipy} library. Subsequently, nonlinear registration to the MNI space was applied to correct for EPI-induced currents. QC was performed during preprocessing by visual inspection of images before and after registration. MRtrix $3.0$ (\url{https://www.mrtrix.org/}) was used to derive an anatomically constrained probabilistic tractography ($2$ million streamlines, step=$0.3 $mm, maximum length=$300$mm, and backtracking) filtered with SIFT2 (\cite{smith2015sift2}). Subject-specific brain parcellations from T1-weighted images were derived using FreeSurfer (\url{https://surfer.nmr.mgh.harvard.edu/}) and used as regions of interest (ROIs) in the SC calculation. The SC matrix was built by counting the number of streamlines connecting all pairs of regions from the Desikan-Killiany (\cite{desikan2006automated}) structural atlas, ignoring self-connections. 
    Similarly to rs-fMRI, each SC matrix was converted into a complete, undirected, and weighted graph. In this case, the edges' weights were defined as the number of streamlines between pairs of ROIs, and the values of the $84$ nodes (i.e., anatomical ROIs) were initialized to one. This SC-based graph was the input to the dMRI channel.

    \subsection{Framework architecture}
    \begin{figure*}[!ht]
        \centering
        \includegraphics[width=\textwidth]{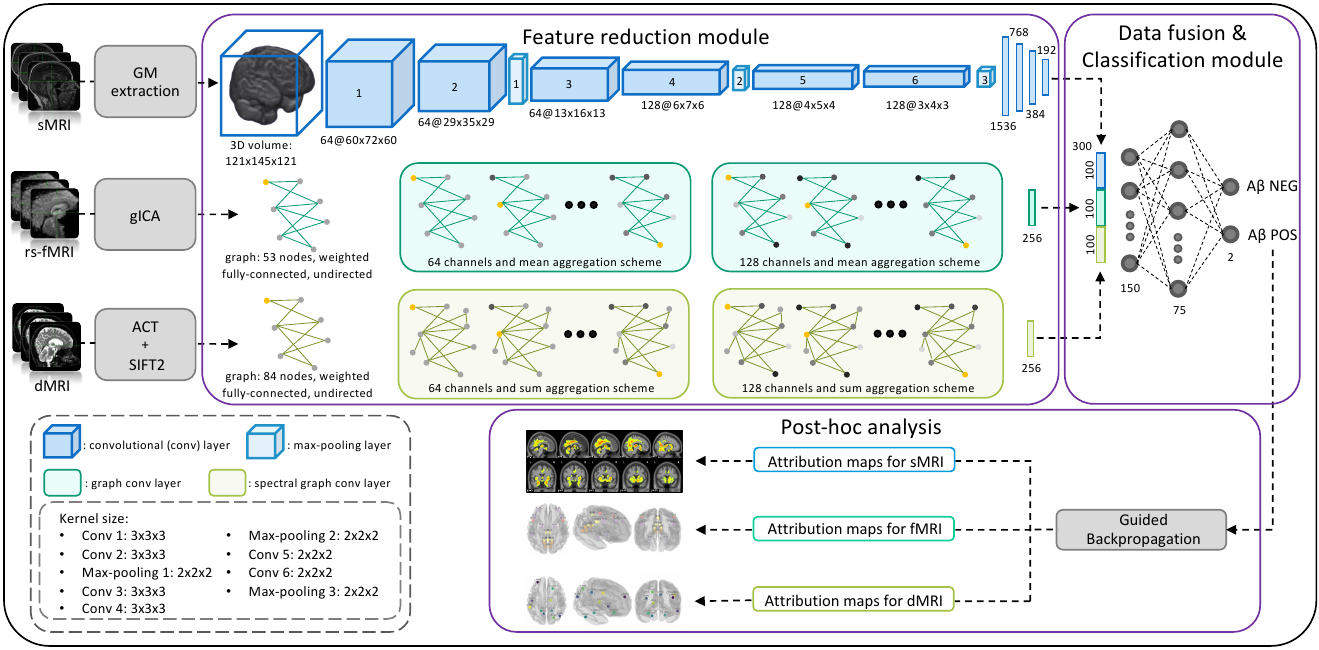}
        \caption{Schematic representation of the proposed framework. The model takes as input three MRI neuroimaging modalities: sMRI 3D volumes, rs-fMRI functional graph, and dMRI structural graph. The DL architecture is composed of two modules: i) a feature reduction module, where the input data are transformed in their latent representations; and ii) a data fusion \& Classification module, where the latent feature of each modality are concatenated together and, finally, they are classified using a MLP.}
        \label{fig:framework}
    \end{figure*}
    
    The proposed framework is shown in Figure \ref{fig:framework}. In detail, the DL architecture used for the classification of A$\beta$ status (positive/negative) has two modules: i) a \textit{feature reduction module} that actuates feature reduction using three different Neural Networks (NNs) to transform the input data into corresponding latent representations; and ii) a \textit{data fusion \& classification module} that concatenates the latent representations of each modality and uses a Multilayer Perceptron (MLP) that takes the fused latent features as input for the final classification. Finally, a \textit{post-hoc} explainability analysis was performed on the correctly classified subjects to highlight the feature contributions to the classification task. In the following paragraphs, the two modules are further described.

        \paragraph{Feature reduction module}
        Three different NNs form the \textit{feature reduction module}. Each NN extracts a latent representation of one modality resulting in a latent vector of $100$ features. The 3D sMRI volumes were analyzed using a 3D-CNN with six convolutional layers and three max-pooling layers, completed with four dense layers.
        Conversely, both FNC and SC graphs were analyzed using two different GCNs. These models update the representation of each node, aggregating neighbor information iteratively in each layer through the message-passing scheme.
        The rs-fMRI channel was analyzed using a GCN with two graph convolutional layers proposed by \cite{morris2019weisfeiler} that were followed by a dense layer. This type of convolutional layer is a powerful operator that integrates both high- and low-level structures along with their relationships into a single graph (\cite{morris2019weisfeiler}).
        The dMRI channel was analyzed with a different GCN with two Chebyshev spectral graph convolutional layers proposed in \cite{defferrard2016convolutional} followed by a dense layer. Spectral convolutional layers are high-performance layers that have been used effectively on irregular graphs (\cite{parisot2018disease}).
        These two architectures were chosen on an empirical basis as those leading to the best unimodal classification accuracy.
        
        \paragraph{Data fusion \& Classification module}
        The \textit{data fusion \& classification module} consists of a fusion layer and a classifier. The fusion layer concatenates the latent features extracted from the three channels, resulting in a vector of $300$ features that incorporate information from all three modalities for each subject. Lastly, the latent vector was used as input for the final MLP classifier, composed of three dense layers.
        The convolutional and dense layers in this framework used a ReLU activation function, except for the last layer, which used a softmax activation function to obtain the classification probabilities for each class.
        
    \subsection{Training scheme and evaluation}
    The model was trained with stratified $5$-fold cross validation on the entire cohort to investigate its generalizability across individuals, and a hyperparameter search was performed empirically through a grid search procedure to maximize the average validation accuracy. To this end, different combinations were tested for the hyperparameters (batch size, learning rate, number of epochs, and regularization parameter). Additionally, different numbers of hidden layers for CNN, GCNs, and classifier were considered, also changing the number of channels for the convolutional layers and the number of neurons in the dense layers. The mini-batch strategy (with $16$ subjects per batch) was finally adopted. The Adam optimizer (learning rate: $0.00001$) was used to update the entire multimodal architecture. L$2$ regularization (weight decay: $0.0001$) was applied to reduce overfitting. Weighted cross entropy was used as the loss function. The model was trained for $200$ epochs.

    Performance was evaluated using the mean evaluation accuracy, precision, recall, and F1 score over the five folds.

    For the sake of comparison, we also tested the unimodal models with the corresponding MRI data, where the architectures were the same as the different branches of the multimodal framework (i.e., 3D-CNN for sMRI, and GCN for both rs-fMRI and dMRI).
    
    \section{Post-hoc analysis}
    \subsection{Guided backpropagation}
    The \textit{post-hoc} XAI analysis was conducted using guided backpropagation (GBP) (\cite{springenberg2014striving}). GBP uses the model gradients to extract the feature contribution maps with the same shape as the input data. It belongs to the "modified backpropagation" class of XAI methods in which the backward flow of gradients is modified with ReLU activation (\cite{rahman2023looking}), setting the negative gradients to zero and only allowing non-negative gradients to be backpropagated. This approach enables the visualization of which input features activated the neurons and most contributed to the final prediction.
    
    \subsection{Contribution maps and Statistical analysis}
    The attribution maps were extracted for the correctly classified A$\beta$ positive subjects. The average A$\beta$ positive subject attribution map was derived for identifying the most important features.
    To evaluate the sMRI GBP contribution maps, the Harvard-Oxford (\cite{desikan2006automated}) and the probabilistic cerebellar (\cite{diedrichsen2009probabilistic}) atlases from FSL were employed to define $56$ different ROIs, including cortical, subcortical, and cerebellum regions. The sum of GBP attributions inside each ROI was calculated for the sMRI and weighted to account for the volume of each specific region. Conversely, the GBP attribution of each node for both rs-fMRI and dMRI was extracted directly from the two GCNs. Then, considering the average map for each modality, the percentage of explanation for each region/node was computed over the total contribution within and across modalities. 
    The top $10$ ROIs (sMRI) and nodes (rs-fMRI and dMRI) with the highest percentage of GBP contribution were selected for further investigation.
    
    Subsequent statistical analyses were performed on the original data for all correctly classified subjects. For the sMRI, the mean values of the top $10$ ROIs resulting from the XAI analysis were extracted from the input GM volumes and used as features for the statistical analysis. For the rs-fMRI and dMRI, graph-based measures were first derived from the full connectivity matrices in order to have a summary measure per node, and then only the top $10$ nodes were retained for both rs-fMRI and dMRI for statistical analysis. In particular, the node strength was computed for the FNC matrices. This is defined as the sum of the weights of the edges connected with a given node, where in absolute terms, higher values mean more important nodes. Betweenness centrality was calculated for SC matrices, representing the fraction of all shortest paths in the SC matrix that contains the node under analysis, where the shortest path is the shortest sequence of nodes between node \textit{i} and node \textit{j}. As the sparsity of the SC matrix could limit the interpretation of the node strength results for the dMRI data, we preferred to rely on a centrality measure for this analysis. In this case, nodes that belong to more paths likely play a pivotal role in the propagation of the information inside the network. These two metrics were computed using the Brain Connectivity Toolbox (BCT) (\cite{rubinov2010complex}) in Matlab.
    
    Mann-Whitney tests were then used to compare the values of all these features for the top $10$ ROIs/nodes between A$\beta$ positive and negative correctly classified individuals.
    Finally, FDR correction for multiple comparisons was applied.

\section{Results}

    \subsection{Classification performance}
    The proposed framework for the classification of A$\beta$ status achieved a mean$\pm$std accuracy, precision, recall, and F1 score of $0.762\pm0.04$, $0.694\pm0.05$, $0.774\pm0.10$, and $0.727\pm0.05$, respectively, across the evaluation folds. 
    
    The single networks that composed the multimodal framework along with the corresponding input data were also tested in the same classification task. Table \ref{tab:ours_classification_models} shows the performance comparisons for the multimodal and the three unimodal models. Results highlight how the multimodal pipeline is able to outperform the unimodal models in terms of accuracy, recall, and F1 score in the same classification task.

    \begin{table}[!ht]
        \centering
        \begin{adjustbox}{width=0.65\linewidth}
        \begin{tabular}{c c c c c}
            \hline
            \makecell{Model} & \makecell{ACC} & \makecell{PRE} & \makecell{REC} & \makecell{F1} \\
            \hline
            Multimodal & \boldmath{$0.762\pm0.04$} & $0.694\pm0.05$ & \boldmath{$0.774\pm0.10$} & \boldmath{$0.727\pm0.05$} \\
            Unimodal sMRI & $0.750\pm0.06$ & \boldmath{$0.721\pm0.06$} & $0.672\pm0.16$ & $0.683\pm0.09$ \\
            Unimodal fMRI & $0.593\pm0.05$ & $0.423\pm0.23$ & $0.303\pm0.21$ & $0.338\pm0.19$ \\
            Unimodal dMRI & $0.603\pm0.04$ & $0.419\pm0.25$ & $0.311\pm0.28$ & $0.322\pm0.23$ \\
            \hline
        \end{tabular}
        \end{adjustbox}
        \caption{Classification performance of the proposed multimodal framework with respect to the unimodal models for sMRI, rs-fMRI, and dMRI for the same classification task. ACC=accuracy, REC=recall, PRE=precision.}
        \label{tab:ours_classification_models}
    \end{table}

    \subsection{GBP-based attribution maps}
    The evaluation set in the fold with the highest evaluation accuracy was used for \textit{post-hoc} analysis. Importantly, in the analyzed fold, the evaluation and training sets were disjoint.
    
\begin{figure*}[!ht]
       \centering
        \includegraphics[width=\textwidth]{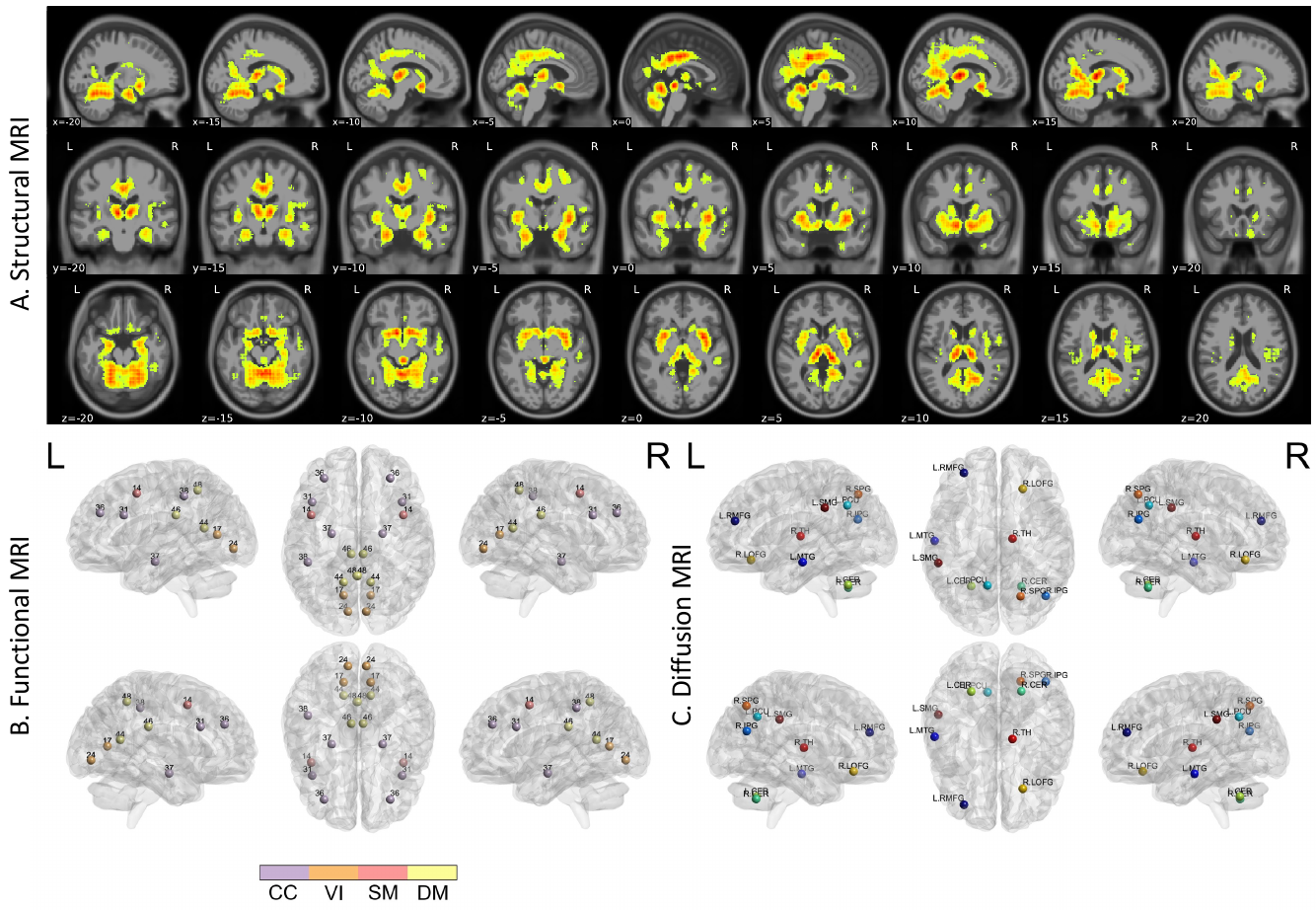}
        \caption{GBP-based attributions for the A$\beta$ positive mean subject derived from the correctly classified individuals overlaid to the MNI152 template, where: A. Saggital, coronal, and axial views for the average sMRI GBP map where only the attributions exceeding the $96^{th}$ percentile are shown, highlighting both cortical and subcortical regions; B. The $10$ most important nodes (ICs) from the rs-fMRI data, representing mainly the DM and CC brain networks; C. The $10$ most important nodes (ROIs) from the dMRI data, involving both cortical and subcortical regions in both hemispheres, also including the cerebellum.}
        \label{fig:results_gbp}
    \end{figure*}

    Figure \ref{fig:results_gbp}A shows the sMRI GBP attribution map for the A$\beta$ mean positive subject, overlaid to the MNI template. Qualitatively, mainly subcortical regions (e.g., \textit{hippocampus}, \textit{thalamus}) were relevant to the final classification along with a few cortical areas.

    Figure \ref{fig:results_gbp}B shows the $10$ most important nodes for the rs-fMRI modality. All displayed nodes (ICs) are bihemispheric, except for the left inferior parietal lobule (IC38) node, which is mainly in the left hemisphere. The most important nodes belong to the \textit{DM} ($3$ nodes), \textit{CC} ($4$ nodes), \textit{VI} ($2$ nodes), and \textit{SM} ($1$ node) networks.

    Finally, Figure \ref{fig:results_gbp}C shows the $10$ most important dMRI nodes (ROIs). Subcortical regions (\textit{thalamus}, \textit{cerebellum}) along with areas in the \textit{frontal}, \textit{temporal}, and \textit{parietal lobes} were identified.

    \subsection{Feature relevance \& Statistical analysis}
    \begin{figure}[!ht]
        \centering
        \includegraphics[width=0.92\linewidth]{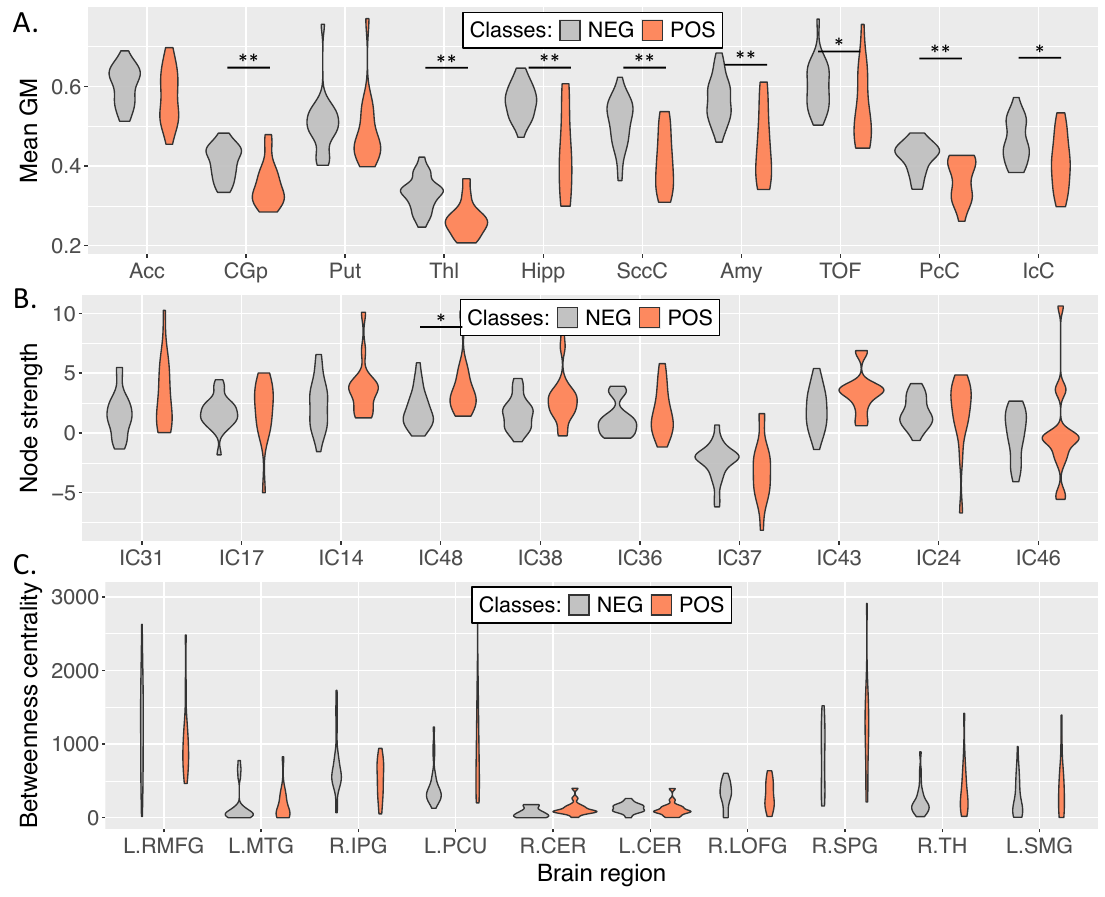}
        \caption{Distributions of the input data for the 10 most important brain regions/nodes considered in the statistical analysis for each modality. A. Between-subject distribution of the regional (mean) GM volumes for sMRI; B. Between-subjects distribution of the node strength values for rs-fMRI; and C. Between-subject distribution of the betweenness centrality values for dMRI.}
        \label{fig:input_feat}
    \end{figure}
    
    Figure \ref{fig:input_feat}A shows the violinplots for the top $10$ sMRI features, while Table \ref{tab:stat_analysis}a shows the corresponding percentages of mean GBP contributions across A$\beta$ positive correctly classified subjects for the same regions, weighted by their volumes.
    The Table also reports the \textit{p}-values and FDR-corrected \textit{p}-values of the statistical tests performed on the input features. 
    We identified significant differences between classes after FDR correction in the \textit{precuneus}, \textit{cingulate gyrus} (posterior division), \textit{thalamus}, \textit{hippocampus}, \textit{supracalcarine and intracalcarine cortices}, \textit{amygdala}, and \textit{temporal occipital fusiform cortex} showing an overall comparison direction of $NEG > POS$.

    Figure \ref{fig:input_feat}B shows violinplots representing the graph node strength for the top $10$ rs-fMRI nodes, while Table \ref{tab:stat_analysis}b shows the mean percentages of GBP contributions extracted from the same nodes (ICs) across A$\beta$ positive patients. After FDR correction, the \textit{precuneus} (IC48) was the only region with significant differences ($NEG < POS$). Before correction, \textit{precuneus} component (IC43) and \textit{precentral gyrus} (IC14) were also significant with $NEG < POS$, while the \textit{left inferior parietal lobule} (IC38) exhibited a trend towards uncorrected statistical significance (\textit{p}$<0.10$, with $NEG < POS$).

    Figure \ref{fig:input_feat}C shows the violinplots representing the graph betweenness centrality for the top $10$ dMRI nodes, and Table \ref{tab:stat_analysis}c shows the mean percentages of GBP contributions for the same nodes across A$\beta$ positive correctly classified subject. The \textit{precuneus LH} and \textit{cerebellum RH} were significant when considering uncorrected \textit{p}-values ($NEG < POS$), but none of the results survived multiple comparisons correction. Moreover, the \textit{right superior parietal gyrus} showed a trend towards significance (\textit{p}$=0.07825$, $NEG < POS$).

    We also computed the percentage of contribution of each region for the three modalities combined, retrieving the $10$ most important areas (marked with $^\dagger$ in Table \ref{tab:stat_analysis}a,b,c). 
    All the $10$ most important brain regions belonged to the sMRI modality (Table \ref{tab:stat_analysis}a).

\section{Discussion}

    In this work, we proposed a multimodal data fusion framework that integrates multiple MRI techniques (sMRI, rs-fMRI, and dMRI) for the classification of A$\beta$ status in an unbalanced cohort of subjects. A \textit{post-hoc} XAI analysis and evaluation was performed using GBP to assess feature importance, complemented by statistical analyses on the so-identified input features for validation via \textit{post-hoc} plausibility assessment.
    
    \begin{table}[!htp]
        \centering
            \centering a) Structural MRI\\[3pt]
            \begin{adjustbox}{width=0.5\linewidth,center}
            \begin{tabular}{l c l l c}
                \hline
                \makecell{Region} & \makecell{Percentage (\%)} & \makecell{p-value} & \makecell{FDR adj.\\p-value} & \makecell{Comparison\\direction} \\
                \hline
                Accumbens$^\dagger$ (Acc)                           & $6.59\%$ & $0.20402$ & $0.22669$ & n.s. \\
                Cingulate gyrus, posterior$^\dagger$ (CGp)          & $4.91\%$ & $\textbf{0.00182}$ & $\textbf{0.00304}$ & NEG$>$POS \\
                Putamen$^\dagger$ (Put)                             & $4.59\%$ & $0.26024$ & $0.26024$ & n.s. \\
                Thalamus$^\dagger$ (Thl)                            & $4.09\%$ & $\textbf{0.00016}$ & $\textbf{0.00146}$ & NEG$>$POS \\
                Hippocampus$^\dagger$ (Hipp)                        & $3.95\%$ & $\textbf{0.00073}$ & $\textbf{0.00146}$ & NEG$>$POS \\
                Supracalcarine cortex$^\dagger$ (SccC)              & $3.81\%$ & $\textbf{0.00043}$ & $\textbf{0.00146}$ & NEG$>$POS \\
                Amygdala$^\dagger$ (Amy)                            & $3.45\%$ & $\textbf{0.00066}$ & $\textbf{0.00146}$ & NEG$>$POS \\
                Temporal occipital fusiform cortex$^\dagger$ (TOF)  & $3.22\%$ & $\textbf{0.01793}$ & $\textbf{0.02561}$ & NEG$>$POS \\
                Precuneus cortex$^\dagger$ (PcC)                    & $3.07\%$ & $\textbf{0.00066}$ & $\textbf{0.00146}$ & NEG$>$POS \\
                Intracalcarine cortex$^\dagger$ (IcC)                         & $2.83\%$ & $\textbf{0.03509}$ & $\textbf{0.04386}$ & NEG$>$POS \\
                \hline\\
            \end{tabular}
            \end{adjustbox}
            \centering b) Resting-state functional MRI\\[3pt]
            \begin{adjustbox}{width=0.5\linewidth,center}
            \begin{tabular}{l c c l l c}
                \hline
                \makecell{Region} & \makecell{RSN} & \makecell{Percentage (\%)} & \makecell{p-value} & \makecell{FDR adj.\\p-value} & \makecell{Comparison\\direction} \\
                \hline
                Middle frontal gyrus$^*$ (IC31)& CC & $9.58\%$ & $0.10597$ & $0.21194$ & NEG$<$POS \\
                Calcarine gyrus (IC17) & VI & $6.79\%$ & $0.34078$ & $0.42597$ & n.s. \\
                Precentral gyrus (IC14) & SM & $6.31\%$ & $\textbf{0.04965}$ & $0.16549$ & NEG$<$POS \\
                Precuneus$^*$ (IC48) & DM & $6.08\%$ & $\textbf{0.00389}$ & $\textbf{0.03892}$ & NEG$<$POS \\
                Left inferior parietal lobule (IC38) & CC & $4.20\%$ & $0.07349$ & $0.18372$ & NEG$<$POS \\
                Middle frontal gyrus$^*$ (IC36) & CC & $4.01\%$ & $0.32634$ & $0.42597$ & n.s. \\
                Hippocampus (IC37) & CC & $3.76\%$ & $0.23658$ & $0.38431$ & n.s. \\
                Precuneus$^*$ (IC43) & DM & $3.65\%$ & $\textbf{0.02623}$ & $0.13115$ & NEG$<$POS \\
                Lingual gyrus (IC24) & VI & $3.23\%$ & $0.38648$ & $0.42942$ & n.s. \\
                Posterior cingulate cortex (IC46) & DM & $2.94\%$ & $0.50672$ & $0.50672$ & n.s. \\
                \hline\\
            \end{tabular}
            \end{adjustbox}
            \centering c) Diffusion MRI\\[3pt]
            \begin{adjustbox}{width=0.5\linewidth,center}
            \begin{tabular}{l c l l c}
                \hline
                \makecell{Region} & \makecell{Percentage (\%)} & \makecell{p-value} & \makecell{FDR adj.\\p-value} & \makecell{Comparison\\direction} \\
                \hline
                Rostral middle frontal LH (L.RMFG) & $39.14\%$ & $0.35559$ & $0.44448$ & n.s. \\
                Middle temporal LH (L.MTG) & $9.28\%$ & $0.49705$ & $0.55227$ & n.s. \\
                Inferior parietal RH (R.IPG) & $6.55\%$ & $0.20919$ & $0.34865$ & n.s. \\
                Precuneus LH (L.PCU) & $5.46\%$ & $\textbf{0.01470}$ & $0.14700$ & NEG$<$POS \\
                Cerebellum RH (R.CER) & $4.48\%$ & $\textbf{0.03887}$ & $0.19433$ & NEG$<$POS \\
                Cerebellum LH (L.CER) & $3.94\%$ & $0.14061$ & $0.28187$ & n.s. \\
                Lateral orbito frontal RH (R.LOFG) & $3.78\%$ & $0.96545$ & $0.96545$ & n.s. \\
                Superior parietal RH (R.SPG) & $3.76\%$ & $0.07825$ & $0.26084$ & NEG$<$POS \\
                Thalamus RH (R.TH) & $3.30\%$ & $0.14094$ & $0.28187$ & n.s. \\
                Supra marginal LH (L.SMG) & $3.29\%$ & $0.31926$ & $0.44448$ & n.s. \\
                \hline
            \end{tabular}
            \end{adjustbox}
            \caption{Percentage of GBP explanations (for the A$\beta$ mean positive subject) and the results from the statistical tests for the top $10$ brain regions derived from: a) sMRI, b) rs-fMRI, and c) dMRI.\\
            The term n.s. means not statistically significant, the sign
            $^\dagger$ means top $10$ regions across the three modalities, while the sign $^*$ means ICs with different spatial locations in rs-fMRI, and LH/RH means left/right hemisphere, respectively, in dMRI.}
            \label{tab:stat_analysis}
    \end{table}

    \subsection{Classification performance}
    
    Results showed that the multimodal framework outperformed single-modality models in terms of classification performance, in particular for rs-fMRI and dMRI, while the sMRI achieved performance close to the multimodal one, but with higher variance across the evaluation folds. This provides evidence of the added value brought by multimodality approaches in terms of classification accuracy while injecting complementary information shading light on the underlying neurophysiological mechanisms, which is particularly relevant for the study of complex neurodegenerative diseases influenced by multi-domain factors.
    
    Table \ref{tab:classification_comparison} shows the performance of state-of-the-art (SOA) works for the classification of A$\beta$ positive versus negative subjects. Of note, current SOA works use multiple datasets and approaches for addressing this task. It goes without saying that the lack of a common reference dataset inherently limits the relevance of performance comparison. However, reaching good accuracy and outperforming the SOA does not exhaust the contribution of the proposed approach, whose potential lies in decoding the neurophysiological changes induced by the A$\beta$ status as captured by the considered MRI modalities.
    
    Most of the SOA works relied on PET scans to address this task, e.g., \cite{kim2021deep} developed a CNN model that takes the different views (axial, coronal, and sagittal) of a 3D volume as input, while \cite{lee2021performance} used three different well-known 3D-CNN (Inception3D, ResNet3D, and VGG3D) models to address this task using the full 3D volume of PET scans. They achieved an accuracy of around $0.710$ on average in the test set, with a maximum accuracy of $0.870$ by \cite{lee2021performance} using the VGG3D architecture.
    In the same way, \cite{ladefoged2023estimation} employed a 3D-CNN to identify the A$\beta$ status reporting both the average validation accuracy and also the accuracy on a hold-out test set (ADNI data) using the ensemble method created with the best models across the different folds, achieving accuracies of around $0.980$. Recently, \cite{rasi2024predicting} employed different combinations of feature selection methods and classifiers in order to analyze FDG-PET-derived features. They achieved an AUC of $0.924$ employing LASSO with the GNB model.
    Only a few works used MRI modalities to classify A$\beta$ positive and negative subjects. \cite{yang2021deep} focused on the hippocampus region only for detecting A$\beta$ positivity, testing the network on two different classification tasks: AD A$\beta$ positive versus CN A$\beta$ negative, and MCI A$\beta$ positive versus MCI A$\beta$ negative, achieving an accuracy of $0.772\pm0.03$ and $0.592\pm0.05$, in the first and second tasks, respectively.
    Lastly, \cite{chattopadhyay2023predicting} implemented a 3D-CNN for detecting A$\beta$ status from sMRI images defining the A$\beta$ status of the CN, MCI, and AD patients considering the PET cortical standardized uptake value ratio reaching an accuracy of $0.760$.
    
    Comparing the proposed method with these six works, we outperformed the method that relied on 3D FDG-PET proposed by \cite{kim2021deep} and the performance of \cite{chattopadhyay2023predicting}. Our study achieved competitive performance in the validation set compared to \cite{lee2021performance} using amyloid PET images, although our pipeline led to higher accuracy compared to the ResNet3D they used, and a similar accuracy to that in \cite{yang2021deep} using the hippocampus region only. 
    Besides providing competitive performance, our model allows to successfully integrate structural GM features extracted from a 3D-CNN along with functional and structural connectivity relying on two {\em ad-hoc} GCNs. Notably, while a few studies have recently started to explore SC/FC in conjunction with GCNs in the AD classification task (\cite{zhang2023multi,zhang2023multirelation}), results are still limited, calling for further investigation in the AD continuum.
    
    \begin{table*}[!htp]
        \centering
        \begin{adjustbox}{width=\linewidth}
            \begin{tabular}{ l  l  l  l  c  l  c  c  c}
                \hline
                \makecell[c]{Authors} & \makecell[c]{Modalities} & \makecell[c]{Study cohort} & \makecell[c]{Input data} &\makecell[c]{Model} & \makecell[c]{ACC} & \makecell[c]{REC} & \makecell[c]{PRE} & \makecell[c]{F1} \\
                \hline
                \cite{kim2021deep} & \makecell{A$\beta$/FDG-PET} & \makecell{$738$ POS, $815$ NEG} & \makecell{3D FDG-PET} & \makecell{2.5-D CNN} & \makecell{$0.733^*$- $0.690^\dagger$} & \makecell{$0.678^*$-$0.768^\dagger$} & \makecell{n.d.} & \makecell{$0.709^*$-$0.712^\dagger$} \\[5pt]
                \cite{lee2021performance} & \makecell{A$\beta$ PET} & \makecell{$350$ POS, $333$ NEG} & \makecell{3D A$\beta$ PET} & \makecell{Inception3D \\ ResNet3D \\ VGG3D} & \makecell{$0.954^*$-$0.767^\dagger$ \\ $0.920^*$-$0.671^\dagger$ \\ $0.977^*$-$0.870^\dagger$} & \makecell{$0.918^*$-$0.845^\dagger$ \\ $0.918^*$-$0.944^\dagger$ \\ $0.959^*$-$0.831^\dagger$} & \makecell{n.d.} & \makecell{n.d.} \\[15pt]
                \cite{ladefoged2023estimation} & \makecell{A$\beta$ PET} & \makecell{POS, NEG n.d.\\$1309+224$} & \makecell{3D A$\beta$ PET} & \makecell{3D-CNN} & \makecell{$0.980^*$-$0.990^\dagger$} & \makecell{$0.980^*$-$0.990^\dagger$} & \makecell{n.d.} & \makecell{$0.980^*$-$0.990^\dagger$} \\[15pt]
                \cite{rasi2024predicting} & \makecell{FDG-PET} & \makecell{$185$ POS, $116$ NEG} & \makecell{FDG-PET-derived\\ features} & \makecell{LASSO+GNB} & \makecell{$0.924$ (AUC)} & \makecell{n.d.} & \makecell{n.d.} & \makecell{n.d.} \\[15pt]
                \cite{yang2021deep} & \makecell{sMRI} & \makecell{$151$ AD POS, $232$ CN NEG\\$171$ MCI POS, $171$ MCI NEG} & \makecell{Hippocampus\\region} & \makecell{SDF-based NN} & \makecell{$^10.772\pm0.03$\\$^20.592\pm0.05$} & \makecell{n.d.} & \makecell{n.d.} & \makecell{n.d.} \\[15pt]
                \cite{chattopadhyay2023predicting} & \makecell{sMRI} & \makecell{POS, NEG n.d.\\$459$ CN, $67$ MCI, $236$ AD} & \makecell{3D volume} & \makecell{3D-CNN} & \makecell{$0.760^\dagger$} & \makecell{n.d.} & \makecell{n.d.} & \makecell{$0.746$} \\[15pt]
                Proposed framework & \makecell[c]{sMRI, fMRI,\\ dMRI} & \makecell{$133$ POS, $185$ NEG} & \makecell{GM volume \\ FNC graph, SC graph} & \makecell{Multimodal DL\\ model} & \makecell{$0.762\pm0.04^*$}  & \makecell{$0.774\pm0.10^*$} & \makecell{$0.694\pm0.05^*$} & \makecell{$0.727\pm0.05^*$} \\
                \hline
            \end{tabular}
        \end{adjustbox}
        \caption{Comparison of the proposed model with other SOA approaches for the classification of amyloid-$\beta$ positive (POS) versus negative (NEG) conditions. $^1$=AD A$\beta$+ vs CN A$\beta$-; $^2$=MCI A$\beta$+ vs MCI A$\beta$-; $^*$=validation set; $^\dagger$=test set; n.d.=not declared}
        \label{tab:classification_comparison}
    \end{table*}
    
    \subsection{Explainability analysis}
    The \textit{post-hoc} analysis, performed on the correctly classified subjects, consisted of two steps: i) extraction of GBP attribution maps for A$\beta$ mean positive subjects as well as the corresponding percentage of contribution for each brain region, both within and between modalities; ii) statistical analysis on the input features of the $10$ most important regions/nodes identified by GBP. 
    
    The analysis of the sMRI attribution maps revealed that multiple brain regions involved in AD neurodegeneration could be identified. These regions had a higher percentage of contribution relative to the others and belonged to subcortical areas (hippocampus, thalamus, putamen, accumbens, and amygdala) and temporal/occipital/parietal areas (posterior cingulate gyrus, precuneus cortex, supracalcarine and intracalcarine cortices, and the temporal occipital fusiform cortex).
    The hippocampus, in particular, is a well-known biomarker for AD that is subject to high levels of atrophy. Some studies have suggested that this atrophy is attributable to the deposition of A$\beta$ plaques (\cite{cantero2016regional}). An important region that is connected with the hippocampus is the cingulate gyrus which showed a strong reduction of GM in AD patients (\cite{green2023increased,jones2006differential}). In a clinical study performed by \cite{kang2021amyloid}, significant associations were detected between A$\beta$ accumulation and GM atrophy in the hippocampus and posterior cingulate gyrus for the MCI and AD A$\beta$ positive subjects. Some works also linked the nucleus accumbens region to AD pathology and progression. Evidence of GM loss and alteration of nucleus accumbens in MCI and AD, with respect to the CN, had previously been highlighted (\cite{yi2016relation,nie2017subregional}). Additionally, \cite{guo2022amyloid} showed how A$\beta$ oligomers in nucleus accumbens can promote synaptic loss and motivation deficits in AD.
    Other subcortical structures, like the thalamus and putamen, have a high atrophy rate in clinical patients relative to healthy individuals (\cite{de2008strongly,kang2021amyloid}), probably due to A$\beta$ deposition. A previous study highlighted an increase in the standardized uptake value ratio (derived from florbetapir PET) in both putamen and thalamus in the preclinical stages of AD (\cite{edmonds2016patterns}).
    Following the statistical analysis performed on the input GM volumes, significant differences were detected in the hippocampus, posterior cingulate gyrus, amygdala, thalamus, and precuneus, highlighting a considerable decrease of GM in the A$\beta$ positive subjects. These findings were consistent with preexisting literature. 
    Additionally, \cite{thal2002phases} highlighted other regions (that we also found to be relevant) involved in the different phases of A$\beta$ deposition (i.e., proisocortex, allocortical areas, diencephalic nuclei, and striatum).
    
    For the rs-fMRI, the brain regions (ICs) that most contributed to the final classification resided in the DM and CC networks, along with three regions for VI and SM networks (two and one, respectively). The DM network is involved in memory, self-knowledge, and thinking, and it is highly related to AD (\cite{pini2021breakdown,green2023increased}). Previous studies also suggested that DMN is associated with the presence of A$\beta$ plaques (\cite{matthews2013brain}). The CC network is associated with selective attention, working memory, and stimulus-response mapping (\cite{sendi2023disrupted,miller2000prefontral}).
    The statistical analysis, performed on the node strength values, detected one region with a significant difference between the two groups after FDR correction and three regions before (uncorrected \textit{p}-values). One more region, the left inferior parietal lobule, exhibited a trend toward significance.
    Specifically, the A$\beta$ positive class had a significantly increased node strength in the precentral gyrus (IC14, SM network), which was consistent with the findings of \cite{guzman2022amyloid} and \cite{duan2017differences}. The precuneus regions (IC43 and IC48, DM network) were also significantly different between groups with increased strength in the positive group. As in other works (\cite{celone2006alterations,dadario2023functional}), this suggested that the precuneus plays a key role and undergoes pathological changes related to AD. Previous works also identified the posterior DM (the precuneus and posterior cingulate cortex) (\cite{palmqvist2017earliest}) and parts of the CC network (i.e., the middle frontal gyrus and hippocampus) (\cite{canuet2015network,pereira2019amyloid}) as strongly affected by A$\beta$ deposition.
    
    Lastly, in the dMRI channel, GBP identified multiple cortical and subcortical regions as relevant to the outcome of our model. The cerebellum resulted to be important for the model. Of note, the rostral middle frontal gyrus was assigned a high percentage of contribution relative to the other regions. Additionally, the middle temporal gyrus, precuneus, thalamus, and superior parietal regions showed high importance relative to the other $84$ regions. After FDR correction, no region had statistically significant group differences in betweenness centrality values. On the other hand, based upon uncorrected \textit{p}-values, significant differences (\textit{p}$<0.05$) were detected in the precuneus and cerebellum (right hemisphere), while the superior parietal region was close to significance (\textit{p}=$0.07825$). 
    Regarding the two nodes with significant uncorrected differences, the precuneus, which is also identified in the rs-fMRI, is an important hub for functional operations because it is highly connected with other regions by both short- and long-range WM fibers (\cite{dadario2023functional}). On the other hand, different studies highlighted how the cerebellum is subject to increase A$\beta$ deposition in AD pathology (\cite{braak1989alzheimer, catafau2016cerebellar}) at different stages (stage $3$ in \cite{braak1998evolution} and stage $5$ in \cite{thal2002phases}) relative to normally aged subjects.
    Widespread structural alterations in WM tracts connecting cortical and subcortical regions in both hemispheres were detected between A$\beta$ negative CN and positive preclinical AD patients by \cite{molinuevo2014white}.
    
    When combining the explanations of all channels, we found that sMRI most contributed to the final prediction. This was expected as we initially assumed that the effects of A$\beta$ plaques would predominantly impact GM volumes.

    \subsection{Common regions across modalities}
    Multiple brain regions were identified as important by GBP across all three channels. Specifically, the \textit{precuneus} was highlighted in all three modalities. It was one of the most interesting regions as it plays a pivotal role in the transmission of functional information due to the high concentration of WM tracts linking the precuneus with other brain areas. Our finding of decreased GM and increased functional node strength for the region in A$\beta$ positive subjects could be related to the functional compensation effects of AD and deserves further investigation (\cite{celone2006alterations,dadario2023functional}). Additionally, the \textit{cingulate gyrus (posterior division)}, \textit{calcarine gyrus}, and \textit{hippocampus} were important in both sMRI and rs-fMRI. Only the \textit{thalamus} was relevant to both sMRI and dMRI, and the \textit{middle frontal gyrus} was the only area important to both rs-fMRI and dMRI. Interestingly, the \textit{middle frontal gyrus} was a region highlighted in both rs-fMRI and dMRI, but not in the sMRI. This further underlines the importance of integrating different (and complementary) modalities to provide a complete picture of the complex pathological mechanisms. 

    \subsection{Main contributions and outcomes}
    We proposed a multimodal and explainable neuroimaging DL model for the classification of amyloid-$\beta$ positive or negative status.
    The main contribution of this work is a framework that i) successfully integrates volumetric features and connectivity information from sMRI, rs-fMRI, and dMRI data, extracted from one 3D-CNN and two GCNs, respectively, ii) obtains competitive classification performance (ACC=$0.762\pm0.04$) relative to the state-of-the-art, iii) provides insight into the brain regions that most contribute to final model outcome using a GBP-based \textit{post-hoc} analysis. Finally, we evaluated the XAI outcomes by assessing the discriminative power of the selected input features across classes.

    With these analyses, we identified multiple brain regions that could be altered by A$\beta$ plaque deposition, resulting in atrophy, functional changes, and WM connectivity changes relative to the $\beta$ negative class. Furthermore, the analysis identified common regions across modalities, including the \textit{precuneus}, \textit{hippocampus}, \textit{thalamus}, \textit{cingulate gyrus}, \textit{calcarine gyrus}, and \textit{middle frontal gyrus}, strengthing the evidence of their involvement in this pathological process. 

    Our findings highlight the utility of MRI for studying the possible effects of A$\beta$ deposition and the importance of integrating complementary information to enable a better understanding of the differences related to amyloid status. 
    However, we acknowledge that the two main limitations of this work are the limited number of subjects in the dataset having both A$\beta$ information and all the considered MRI modalities, and the unbalanced cohort including CN, SMC, and EMCI in the A$\beta$ negative group and EMCI, LMCI, and AD in the A$\beta$ positive one. Hence, the outcomes could be guided by those factors that were influenced by the amyloid accumulation as captured by the specific modality. In consequence, the GM branch captured atrophy, while the structural and functional connectivity branches captured the respective amyloid-induced connectivity modulations, leading to a combined effect due to A$\beta$ and dementia-related processes. Further studies are thus needed to assess the generalizability of the proposed multimodal framework to larger and possibly employing balanced datasets, relying on study cohorts composed of all clinical classes (CN, SMC, EMCI, LMCI, and AD) in both groups analyzed, A$\beta$ negative and positive individuals, in order to disentangle the contributions of the different factors and hence having the outcomes only related to A$\beta$.

\section{Conclusion}

    In this work, we presented a multimodal and explainable DL-based framework for the classification of amyloid-$\beta$ status, exploiting anatomical and connectivity MRI-based information. The application of GBP enabled the identification of the regions most important to the final model predictions, some of which were common across modalities, (e.g., the precuneus, hippocampus, thalamus, cingulate gyrus, calcarine gyrus, and middle frontal gyrus). Our study demonstrates the potential viability of non-invasive MRI-based detection of amyloid-$\beta$ status involving multimodal data, paving the way for further research in this direction.

\section*{acknowledgments}
Data collection and sharing for this project was funded by the Alzheimer's Disease Neuroimaging Initiative (ADNI) (National Institutes of Health Grant U01 AG024904) and DOD ADNI (Department of Defense award number W81XWH-12-2-0012). ADNI is funded by the National Institute on Aging, the National Institute of Biomedical Imaging and Bioengineering, and through generous contributions from the following: AbbVie, Alzheimer’s Association; Alzheimer’s Drug Discovery Foundation; Araclon Biotech; BioClinica, Inc.; Biogen; Bristol-Myers Squibb Company; CereSpir, Inc.; Cogstate; Eisai Inc.; Elan Pharmaceuticals, Inc.; Eli Lilly and Company; EuroImmun; F. Hoffmann-La Roche Ltd and its affiliated company Genentech, Inc.; Fujirebio; GE Healthcare; IXICO Ltd.; Janssen Alzheimer Immunotherapy Research \& Development, LLC.; Johnson \& Johnson Pharmaceutical Research \& Development LLC.; Lumosity; Lundbeck; Merck \& Co., Inc.; Meso Scale Diagnostics, LLC.; NeuroRx Research; Neurotrack Technologies; Novartis Pharmaceuticals Corporation; Pfizer Inc.; Piramal Imaging; Servier; Takeda Pharmaceutical Company; and Transition Therapeutics. The Canadian Institutes of Health Research is providing funds to support ADNI clinical sites in Canada. Private sector contributions are facilitated by the Foundation for the National Institutes of Health (\url{www.fnih.org}). The grantee organization is the Northern California Institute for Research and Education, and the study is coordinated by the Alzheimer’s Therapeutic Research Institute at the University of Southern California. ADNI data are disseminated by the Laboratory for Neuro Imaging at the University of Southern California. 

This study was funded by NIH grant (RF1AG063153) and NSF Grant (2112455), as well as Fondazione CariVerona (EDIPO project, num. 2018.0855.2019) and MIUR D.M. 737/2021 ``AI4Health: empowering neurosciences with eXplainable AI methods".

\section*{CRediT authors contributions statement}
\textbf{Giorgio Dolci:} Conceptualization, Data curation, Formal analysis, Investigation, Methodology, Software, Writing - original draft. 
\textbf{Charles A. Ellis:} Methodology, Writing - original draft.
\textbf{Federica Cruciani:} Methodology, Data curation, Writing - review \& editing.
\textbf{Lorenza Brusini:} Data curation, Writing - review \& editing.
\textbf{Anees Abrol:} Data curation, Writing - review \& editing.
\textbf{Ilaria Boscolo Galazzo:} Investigation, Data curation, Writing - review \& editing.
\textbf{Gloria Menegaz:} Conceptualization, Supervision, Writing - review \& editing, Funding acquisition.
\textbf{Vince D. Calhoun:} Conceptualization, Supervision, Writing - review \& editing, Funding acquisition.

\section*{Competing Interests}
The authors declare that they have no known competing financial interests or personal relationships that could have appeared to influence the work reported in this paper.

\section*{Additional information}
The data used in this work were collected by ADNI (\url{https://adni.loni.usc.edu/}), and they are publically available after requested on the ADNI website.

\section*{Ethical standard}
The data used in this work were acquired by ADNI (\url{https://adni.loni.usc.edu/}). Information regarding the ethical standard and informed consent of ADNI 3 Protocol are available at the following link on the: \url{https://adni.loni.usc.edu/wp-content/themes/freshnews-dev-v2/documents/clinical/ADNI3_Protocol.pdf}.

\bibliographystyle{unsrt}  
\bibliography{main}  

\end{document}